\documentclass[11pt]{article}

\usepackage[preprint]{acl}

\usepackage{times}
\usepackage{latexsym}

\usepackage[T1]{fontenc}

\usepackage[utf8]{inputenc}

\usepackage{microtype}

\usepackage{inconsolata}

\usepackage{graphicx}

\usepackage{booktabs}

\usepackage{subfigure} 

\usepackage{multirow}
\usepackage{multicol}
\usepackage{amssymb} 
\usepackage{colortbl} 
\usepackage[table,xcdraw,dvipsnames]{xcolor} 
\usepackage{amsmath, amssymb, amsthm}
\usepackage{float}
\usepackage{arydshln}
\usepackage{pifont}
\usepackage{enumitem}

\usepackage{makecell}
\usepackage[normalem]{ulem}

\author{
  \textbf{Shuyu Guo}$^{1}$\thanks{Work performed during an internship at Huawei.} \quad
  \textbf{Wenxiang Hu}$^{2}$ \quad
  \textbf{Yuyue Zhao}$^{2}$ \\
  \textbf{Lun Zhou}$^{2}$ \quad
  \textbf{Yougang Lyu}$^{2}$\thanks{Corresponding authors.} \quad
  \textbf{Xiaohui Yan}$^{2}$\footnotemark[2] \quad
  \textbf{Zhaochun Ren}$^{3}$ \\[2pt]
  $^{1}$Shandong University \quad
  $^{2}$Huawei Technologies Co., Ltd. \quad
  $^{3}$Leiden University \\[2pt]
  \texttt{guoshuyu225@gmail.com}, \texttt{huwenxiang3@huawei.com}, \\
  \texttt{\{yuyuezhao, yougang.lyu\}@huawei-partners.com},
  \texttt{zhoulun1@huaiwei.com}\\
  \texttt{yanxiaohui2@huaiwei.com}, \texttt{z.ren@liacs.leidenuniv.nl}
}

\title{RubricReviewer: From Direct Critique to Rubric-Driven Review via Agentic Evidence Synthesis and Human-Aligned Generation}

\begin{document}
\maketitle

\begin{abstract}
Peer review at major venues is under unprecedented submission pressure, motivating the use of large language models (LLMs) as review assistants.
Existing LLM-based reviewers, however, face two structural limitations.
First, they map manuscripts directly to reviews, leaving the underlying rubric implicit and entangling its derivation with the judgement.
Second, the prevailing paradigms each capture only half of a good review: training-free agents gather broad evidence but produce undirected critiques, while training-based reviewers inherit human discriminative judgement together with its noise and uneven coverage.
We introduce \textbf{RubricReviewer}, a fully rubric-driven framework that addresses both limitations.
It makes rubric generation an explicit intermediate step, so that both review generation and the final assessment are conditioned on paper-adaptive rubrics.
It further combines a training-free agent (\textbf{Scout}) that gathers external evidence with a human-aligned trained model (\textbf{Aligner}) that consumes this evidence, fusing the strengths of both supervision sources.
Experiments on real-world submissions show that \textbf{RubricReviewer} produces reviews that are markedly more comprehensive and more discriminative than prior systems, and exhibits the strongest robustness against adversarial prompt-injection attacks.
Ablation studies further confirm the necessity of each component.
\end{abstract}
\section{Introduction}

Peer review underpins scientific progress, but its capacity has not kept pace with the explosive growth of submissions~\cite{llmpeerchallenges}.
A reliable automated peer-review system would relieve reviewers from heavy workload while providing authors with actionable, paper-specific feedback to accelerate their manuscript refinement.

Automated peer review has accordingly attracted growing attention.
Early work framed the problem as score regression or accept/reject classification on static datasets~\cite{peerread,nlpeer}.
With the advent of large language models (LLMs), the focus has shifted toward generating the textual review directly.
Current methods broadly cluster into two paradigms.
\emph{Training-based} methods fine-tune LLMs on real human reviews to align outputs with expert judgements~\cite{cycleresearcher,deepreview}.
\emph{Training-free} methods build LLM-agent pipelines that operate without parameter updates, typically through retrieval-augmented evidence gathering~\cite{ReviewGrounder} or multi-agent committee simulation~\cite{aiscientist,agentreview}.

Despite these advances, the prevailing paradigm still suffers from two limitations.
(1) Existing methods map a manuscript directly to its final review, leaving the underlying rubric implicit.
Such direct generation entangles two distinct subprocesses, deriving the rubric and judging the paper against it, thereby substantially raising the learning difficulty.
(2) These two paradigms each capture only half of what a good review requires.
Training-free LLMs excel at broad evidence gathering and summarisation, yet tend to produce untargeted critiques that gravitate toward neutral judgements~\cite{unveildefects,faultyreasoning}.
Training-based reviewers inherit the discriminative judgements humans deliver, yet also their noise and uneven coverage~\cite{goodbadconstructive,beyondrating}.

To address both limitations, we propose \textbf{RubricReviewer}, a framework comprising two key designs, each tackling one limitation.
For (1), we introduce explicit rubric generation as an intermediate step, decomposing peer review into three sequential stages, namely rubric generation, rubric-conditioned review generation, and final assessment.
For (2), our pipeline combines both supervision sources via two cooperating components.
\textbf{Scout} is a training-free set of role-specialised LLM agents that retrieve external evidence and produce structured analyses as references.
\textbf{Aligner} is a trained model that consumes these references and produces outputs aligned with human judgement.
Together, \textbf{RubricReviewer} fuses the broad coverage and evidence-gathering strengths of agentic LLMs with the discriminative judgements distilled from human reviews.

Extensive experiments on real-world submissions demonstrate \textbf{RubricReviewer}'s superior performance across multiple dimensions.
The rubrics underlying our reviews cover human-essential evaluation dimensions far more comprehensively than any baseline, and the resulting per-rubric reviews show substantially higher agreement with human reviewers while being markedly more discriminative.
Building on these more comprehensive and discriminative reviews, the final assessment also attains the best alignment with human ratings and decisions.
Ablation studies confirm the necessity of each design: removing either the rubric-based decomposition or the cooperation between \textbf{Scout} and \textbf{Aligner} causes a consistent performance drop.
\textbf{RubricReviewer} also exhibits strong robustness under adversarial prompt-injection attacks.

With \textbf{RubricReviewer}, we explore a new direction for automated peer review through a fully rubric-driven framework.
Both review generation and the final assessment are conditioned on rubrics dynamically derived from each manuscript, recasting an inherently subjective open-ended task as a collection of objective per-rubric judgements.
It is also an attempt to jointly harvest the complementary benefits of agentic LLMs and supervised fine-tuning on human reviews within a single pipeline.
Its competitive performance points to a promising path toward reliable, comprehensive, and discriminative LLM-based peer reviewers.
\section{Related Work}
\label{sec:related}

\subsection{Rubric-based Evaluation}

Evaluating open-ended tasks without verifiable ground truth has long been challenging~\cite{krishna-etal-2021-hurdles}.
Rubric-based methods address this by decomposing a subjective objective into a set of fine-grained, binary-checkable criteria, recasting holistic quality judgement as a collection of objectively answerable sub-questions~\cite{DBLP:journals/corr/abs-2507-17746,DBLP:journals/corr/abs-2507-18624}.
The resulting structured signal is interpretable and extends supervision to non-verifiable domains, serving both as an evaluation metric~\cite{DBLP:journals/corr/abs-2511-07685,DBLP:journals/corr/abs-2505-08775} and as a reinforcement-learning reward~\cite{DBLP:journals/corr/abs-2509-21500,DBLP:journals/corr/abs-2508-16949}.

Early work adopts predefined rubrics shared across all inputs, which are easy to deploy but lack granularity~\cite{DBLP:conf/acl/HashemiERDK24,DBLP:journals/corr/abs-2409-16191,DBLP:conf/naacl/ShaoJK0KL24}.
Subsequent work moves to query-specific rubrics tailored to each input~\cite{autorubric,openrubrics}, and further to adaptive rubrics that evolve with model behaviour~\cite{DBLP:journals/corr/abs-2602-10885} or align with human preferences during training~\cite{DBLP:journals/corr/abs-2602-03619,DBLP:journals/corr/abs-2602-01511}.
Such instance-aware rubrics have yielded notable gains in long-form report generation~\cite{drtulu}, complex reasoning~\cite{DBLP:journals/corr/abs-2602-10885}, and other open-ended tasks~\cite{DBLP:journals/corr/abs-2505-08775}.

Peer review of scientific manuscripts shares this difficulty.
Reviews are free-form and highly subjective, with no single verifiable answer~\cite{DBLP:journals/corr/abs-2509-09912}.
Yet existing LLM-based peer reviewers map a manuscript directly to a review~\cite{cycleresearcher}, leaving the underlying evaluation rubric implicit and the supervision signal coarse.
The closest prior attempt incorporates rubrics into the review generation process but relies on a small set of meta-rubrics shared across all papers~\cite{ReviewGrounder}.
In contrast, our work drives the peer review pipeline with paper-adaptive rubrics dynamically generated for each manuscript, using them to condition both review generation and final assessment.

\subsection{LLM-based Peer Review}

The growing volume of conference submissions has made efficient and reliable peer review assistance increasingly valuable, both for reducing reviewer burden and for helping authors improve their manuscripts~\cite{llmpeerchallenges,reviewriter,metawriter}.
Early work framed automated peer review as score regression or accept/reject classification on static datasets~\cite{peerread, reviewadvisor,nlpeer}, while more recent efforts shift toward LLM-based review generation that directly produces textual feedback~\cite{autoreviewgen}.

Existing LLM-based peer reviewers fall broadly into two paradigms: training-based and training-free.
Training-based methods fine-tune LLMs on real human reviews to align outputs with human judgments~\cite{openreviewer,cycleresearcher,deepreview}.
Common instantiations include aspect- or template-conditioned supervised fine-tuning~\cite{reviewer2,openreviewer,cycleresearcher}, fine-tuning on long chain-of-thought reasoning traces~\cite{deepreview}, citation-grounded supervised fine-tuning~\cite{echoreview}, and, more recently, reinforcement learning with rating or preference signals~\cite{reviewrl}.
Training-free methods instead build LLM-agent pipelines without parameter updates~\cite{aiscientist,agentreview,marg}.
Representative designs include retrieval- or tool-augmented reviewers that ground critiques in prior literature~\cite{ReviewGrounder}, multi-agent collaborations that simulate the role-based discussion of a review committee~\cite{agentreview,marg,scholarpeer,reviewagents,aiscientist,deepreviewer2}, and question-driven agents that decompose reviewing into iterative checks~\cite{treereview,hgrdebate}.

Despite this progress, generated reviews still fall short of the substantive, well-targeted feedback that human reviewers provide~\cite{liang2024llmfeedback,faultyreasoning}, and the two paradigms exhibit complementary weaknesses.
Training-free agents tend to produce superficial critiques that gravitate toward neutral assessments~\cite{DBLP:journals/corr/abs-2509-09912,unveildefects}, while training-based reviewers inherit the noise and uneven coverage of their human supervision~\cite{goodbadconstructive,beyondrating}.
The two signals are in fact complementary.
LLMs excel at broad evidence gathering and summarisation, allowing them to refine human reviews and broaden the set of evaluation dimensions, whereas human reviews supply targeted judgments that ground the model.
Our work bridges these two sources to generate reviews that are both comprehensive and aligned with human judgment.

\section{Method}
\label{sec:method}

We formalize peer review as a sequential rubric-centric generation process (\S\ref{sec:method-task}), realize it through a multi-stage pipeline that pairs a trained model \textsc{Aligner} with role-specialized LLM agents \textsc{Scout} (\S\ref{sec:method-inference}), and describe how \textsc{Aligner} is trained (\S\ref{sec:method-training}).

\subsection{Task Formulation}
\label{sec:method-task}

\begin{figure*}[ht!]
  \centering
  \includegraphics[width=1.0\textwidth]{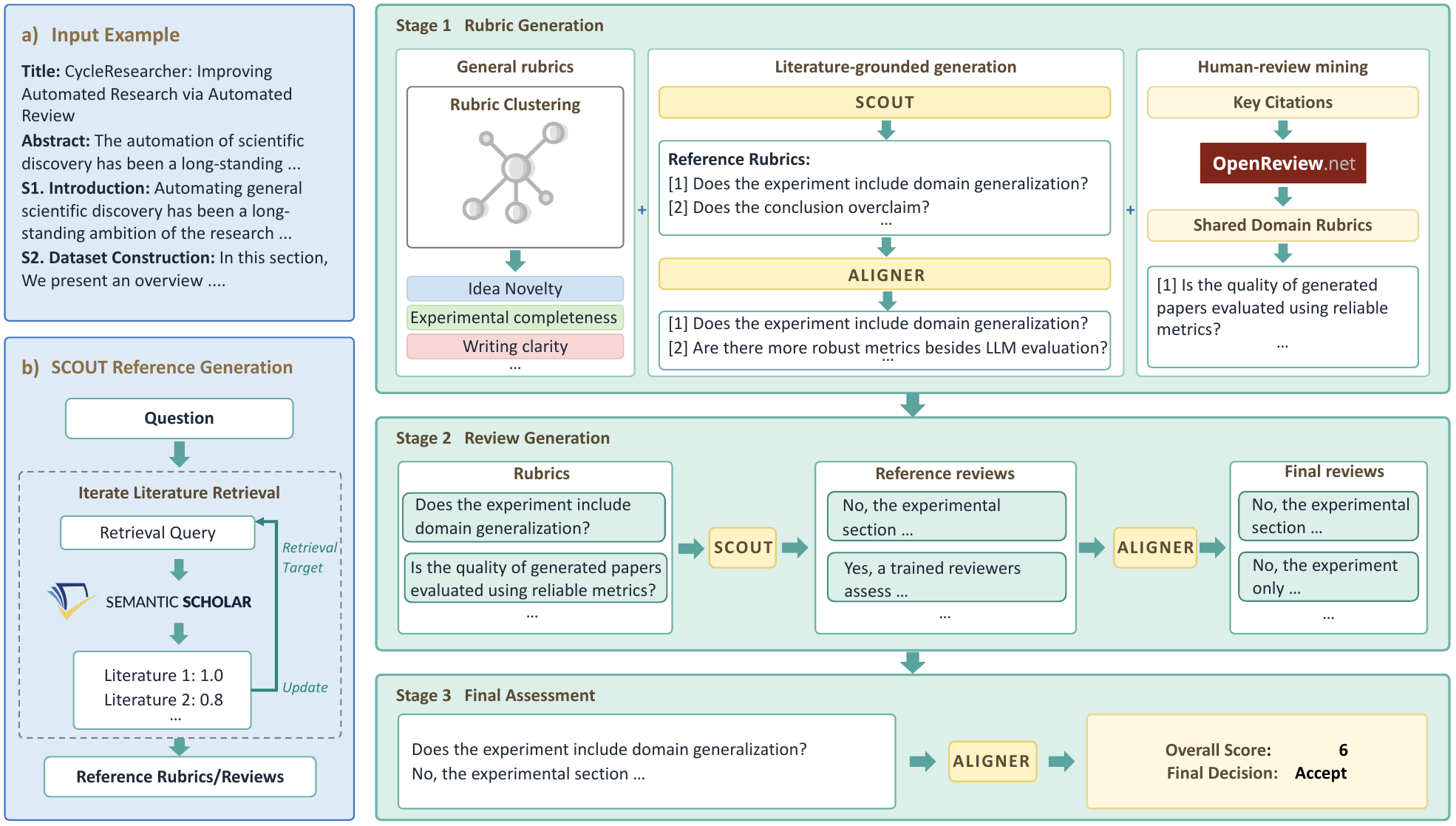}
  \vspace{-3pt}
  \caption{Overview of \textsc{RubricReviewer}'s inference pipeline. Inference unfolds in three stages: rubric generation, review generation, and final assessment. At every stage, \textsc{Scout}, a set of training-free LLM agents, retrieves external literature and produces a structured reference (candidate rubrics or draft reviews), while \textsc{Aligner}, a human-aligned trained model, consumes this reference and emits the final output. In Stage 1, the rubric set fuses three sources: predefined general rubrics, paper-specific rubrics produced by \textsc{Scout}'s literature-grounded generation and then refined by \textsc{Aligner}, and domain-specific rubrics distilled from the public human reviews of key citations.}
  \label{fig:Overview}
\end{figure*}

Formally, given an input manuscript $\mathcal{P}$, automated peer review aims to learn the conditional distribution $p(R, y \mid \mathcal{P})$, where $R = \{r_1, \ldots, r_n\}$ is a set of atomic reviews (each $r_i$ conveys a single evaluative judgment) and $y$ is an optional final assessment such as an overall rating or accept/reject decision.
Every atomic review $r$ implicitly instantiates an underlying evaluation dimension, which we call a rubric and denote $\bar{r}$.
For instance, the atomic review ``\textit{all the proposed components have already appeared in prior work}'' instantiates the rubric ``\textit{is the methodology sufficiently novel relative to prior work?}''.

We decouple rubric generation from review generation by introducing the rubric 
set $\bar{R} = \{\bar{r}_1, \ldots, \bar{r}_K\}$ as a latent intermediate.
Each rubric $\bar{r}_k$ is instantiated by a review subset $R_k \subseteq R$, and all subsets partition the review set $R = \bigsqcup_{k=1}^{K} R_k$.
Peer review is then formulated as the sequential chain $\mathcal{P} \rightarrow \bar{R} \rightarrow R \rightarrow y$, whose joint distribution factorizes as
\begin{equation}
\begin{split}
p(\bar{R}, R, y \mid \mathcal{P}) =\ & p(\bar{R} \mid \mathcal{P}) \, p(R \mid \bar{R}, \mathcal{P}) \\
& \cdot p(y \mid \bar{R}, R, \mathcal{P}),
\end{split}
\label{eq:joint}
\end{equation}
where $p(\bar{R} \mid \mathcal{P})$ generates the rubrics applicable to evaluating the manuscript, $p(R \mid \bar{R}, \mathcal{P})$ generates the atomic reviews instantiating each rubric, and $p(y \mid \bar{R}, R, \mathcal{P})$ generates the final assessment.

This factorization is motivated by a key observation about peer review: although individual atomic reviews vary sharply across papers, the rubrics they instantiate are largely shared.
Direct mapping $\mathcal{P} \rightarrow R$ entangles the paper-independent what with the paper-specific how into a single high-variance generation problem.
Introducing $\bar{R}$ as an intermediate decouples them into two more concentrated subproblems that are each easier to learn.

\subsection{Multi-stage Inference}
\label{sec:method-inference}

To realize the joint factorization in \eqref{eq:joint}, inference proceeds in three stages: rubric generation $p(\bar{R} \mid \mathcal{P})$, review generation $p(R \mid \bar{R}, \mathcal{P})$, and final assessment $p(y \mid \bar{R}, R, \mathcal{P})$.
The pipeline is built around two cooperating components: \textsc{Scout}, a set of LLM agents that autonomously retrieve and summarize external evidence as a reference, and \textsc{Aligner}, a trained model aligned with human judgment that consumes Scout's reference and produces the final output at each stage.

\paragraph{Stage 1: Rubric generation.}
We organize rubrics into three categories: general rubrics $\bar{R}_{\text{general}}$ that apply to all papers (e.g., idea novelty, experimental completeness); domain-specific rubrics $\bar{R}_{\text{domain}}$ shared across all papers in a research area; and paper-specific rubrics $\bar{R}_{\text{paper}}$ tailored to the manuscript at hand.
$\bar{R}_{\text{general}}$ is predefined and contributed in full to the final rubric set.
We defer its construction to \S\ref{sec:method-training}.
$\bar{R}_{\text{domain}}$ and $\bar{R}_{\text{paper}}$ are produced by two parallel paths described below, decomposed into sub-steps $z_1$--$z_5$.

\noindent\textbf{Path 1: literature-grounded generation, producing rubrics in both $\bar{R}_{\text{domain}}$ and $\bar{R}_{\text{paper}}$.}
$(z_1)$ \emph{Key citation identification.} Scout analyzes $\mathcal{P}$ to select its top-$k$ most central references and retrieves their metadata (title, year, abstract) via the Semantic Scholar API, yielding the key citation set $\mathcal{C}^\star$.
$(z_2)$ \emph{Key concern generation.} Genrating appropriate rubrics requires sufficient context about the research area.
Conditioned on $(\mathcal{P}, \mathcal{C}^\star)$, Scout generates a set of key concerns $\mathcal{Q}$ (e.g., dominant paradigms, open problems) whose resolution requires external evidence.
$(z_3)$ \emph{Iterative literature retrieval.} For each concern $q \in \mathcal{Q}$, Scout iteratively queries Semantic Scholar to gather supporting evidence.
The first-round query is generated directly from $q$.
In each subsequent round, Scout refines the query conditioned on the query history, the average relevance of each prior query's results, and the current top-$k$ retrieved pool.
Returned items are scored by relevance to $q$ and merged into the running top-$k$ pool.
Retrieval terminates once the number of above-threshold items reaches a target $K$ or the round budget is exhausted, and the final top-$k$ pool is taken as the evidence pool $\mathcal{L}$.
$(z_4)$ \emph{Reference rubric drafting.} Conditioned on $(\mathcal{P}, \mathcal{Q}, \mathcal{L})$, Scout drafts a candidate rubric set $\bar{R}_{\text{ref}}$.
Taking this draft as a reference, Aligner then generates the final, human-aligned rubrics.

\noindent\textbf{Path 2: human-review mining, producing rubrics in $\bar{R}_{\text{domain}}$.}
Path 2 reuses the key citation set $\mathcal{C}^\star$ from $z_1$ and proceeds with a single further sub-step.
$(z_5)$ \emph{Cross-paper rubric mining.} Since $\mathcal{C}^\star$ largely shares the research area with $\mathcal{P}$, public human reviews of these citations expose key domain-level evaluation dimensions transferable to $\mathcal{P}$.
To leverage this, Scout queries OpenReview for the human reviews of works in $\mathcal{C}^\star$, extracts the rubric items those reviews implicitly invoke, and filters them against $\mathcal{P}$ to retain only those still applicable to the manuscript.

The final rubric set $\bar{R}$ is the deduplicated union $\bar{R}_{\text{general}} \cup \bar{R}_{\text{domain}} \cup \bar{R}_{\text{paper}}$.

\paragraph{Stage 2: Review generation.}
Given $\bar{R}$, atomic reviews are generated rubric by rubric.
For each $\bar{r}_k \in \bar{R}$, Scout first decides whether external evidence is required to judge $\bar{r}_k$. 
If so, Scout generates a retrieval target from $(\mathcal{P}, \bar{r}_k)$ and runs the same iterative literature retrieval as in $z_3$, yielding an evidence pool $\mathcal{L}_k$; otherwise $\mathcal{L}_k = \emptyset$. 
Conditioned on $(\mathcal{P}, \bar{r}_k, \mathcal{L}_k)$, Scout produces a draft judgment $\tilde{R}_k$.
Taking $\tilde{R}_k$ as a reference, Aligner generates the final, human-aligned atomic reviews $R_k$ instantiating $\bar{r}_k$.
The full review is the partition $R = \bigsqcup_{k=1}^{K} R_k$.

\paragraph{Stage 3: Final assessment.}
Aligner consumes the abstract of $\mathcal{P}$ together with all rubric--review pairs $\{(\bar{r}_k, R_k)\}_{k=1}^{K}$ and produces the overall score and accept/reject decision $y$.

\subsection{Model Training}
\label{sec:method-training}

\textsc{Scout} is training-free. We therefore construct three supervised datasets, one per inference stage of \S\ref{sec:method-inference}, to align \textsc{Aligner} with human review behaviour. These are derived from a base corpus $\mathcal{D}$ (DeepReview-13k), where each example provides the paper text $\mathcal{P}$, the set of all reviewers' raw reviews $R$ with their per-reviewer scores, and the final accept/reject decision from the meta-reviewer.

\paragraph{Rubric-generation data.}
$\mathcal{D}$ provides $R$ but not the underlying rubrics $\bar{R}$, so we recover them first. 
Using GPT-5.2, every review in $R$ is decomposed into atomic reviews $\{r_i\}$.
Each $r_i$ is reverse-inferred to its rubric, and atomic reviews sharing the same rubric are aggregated into rubric--review pairs $\{(\bar{r}_k, R_k)\}$.

To build a stable rubric vocabulary across $\mathcal{D}$, we cluster the recovered rubrics. 
Each rubric is tagged with one of three categories \{Idea \& Methodology, Experiment \& Evaluation, Presentation \& Writing\}. 
Within each category, rubrics are sentence-encoded and grouped by coarse-to-fine clustering, and an LLM summarises each cluster into a canonical rubric. 
Canonical rubrics are then iteratively merged.
In each round, an LLM inspects every canonical rubric against its top-$k$ nearest neighbours by embedding centroid and merges those describing the same dimension.
Iteration terminates once the taxonomy stabilises. 
Canonical rubrics covering more than a fraction $\tau$ of $\mathcal{D}$ form the general rubrics ($40$ in total, listed in the Appendix~\ref{sec:general-rubrics}).

For each $\mathcal{P} \in \mathcal{D}$, we run Scout's $z_1$--$z_4$ to obtain its reference rubrics $\bar{R}_{\text{ref}}$. 
The training pair takes $(\mathcal{P}, \bar{R}_{\text{ref}})$ as input and the paper's domain- and paper-specific canonical rubrics as the target.
General rubrics are excluded from the target since they are always added directly at inference.

\paragraph{Review-generation data.}
Pooled from multiple reviewers, raw reviews $R_k$ is often inconsistent across members and uneven in quality. 
We use GPT-5.2 to refine each $R_k$ into a clean target $R_k^\star$. Conditioned on $(\mathcal{P}, \bar{r}_k, R_k)$, GPT-5.2 first decides whether external evidence is needed.
If so, it generates a retrieval target and runs the same iterative retrieval as $z_3$, otherwise the evidence pool is empty. 
The refinement is constrained to re-articulate.
Every claim in $R_k^\star$ must be supported by at least one atomic review in $R_k$, and no new opinions are allowed.

For each $(\mathcal{P}, \bar{r}_k)$, we further run Stage 2's Scout sub-step to obtain its draft judgment $\tilde{R}_k$.
The training pair takes $(\mathcal{P}, \bar{r}_k, \tilde{R}_k)$ as input and $R_k^\star$ as the target.

\paragraph{Final-assessment data.}
$\{(\bar{r}_k, R_k^\star)\}$ from the review-generation data is restricted to rubrics surfaced by raw human reviews, and does not cover the full inference-time rubric set $\bar{R}$ (\S\ref{sec:method-inference}). 
For rubrics in $\bar{R} \setminus \{\bar{r}_k\}$, we generate pseudo-golden reviews via Stage 2 inference with the trained Aligner.

The training pair takes the abstract of $\mathcal{P}$ together with all resulting rubric--review pairs as input and $y = (\bar{s}, d)$ as the target, where $\bar{s}$ is the mean of per-reviewer scores rounded to the nearest integer and $d$ is the meta-reviewer's accept/reject decision.

\section{Experimental Setup}

\subsection{Dataset}
We conduct all experiments on \textbf{DeepReview-13K}~\cite{deepreview}, a curated dataset of ICLR 2024--2025 submissions paired with their human-written reviews and assessments. 
To align this raw resource with our methodological paradigm, we use GPT-5.2 for data rewriting and extension, such as rubric extraction, literature retrieval, and review refinement.
We follow the standard split with approximately 12K papers for training.
To control training and inference cost, the review generation stage only uses a one-quarter subsample and evaluation is restricted to the first 200 papers of the ICLR~2025 test subset.

\subsection{Baselines}
We compare \textsc{RubricReviewer} against three categories of baselines:
\begin{itemize}[leftmargin=1.2em,itemsep=0pt]
\item \textbf{Foundation LLM (zero-shot).} GPT-5.2\footnote{\url{https://openai.com/index/introducing-gpt-5-2/}} prompted directly on the paper without any specialized framework. 
For fair comparison, all direct LLM API calls in our pipeline and the baselines use GPT-5.2.
\item \textbf{Training-free agentic frameworks.} \textsc{AI Scientist}~\cite{aiscientist} and \textsc{AgentReview}~\cite{agentreview}, which simulate peer review through multi-agent prompting, and \textsc{DeepReviewer~2.0}~\cite{deepreviewer2}, an agentic system that produces traceable, auditable reviews via tool-integrated reasoning.
\item \textbf{Training-based reviewer models.} \textsc{CycleReviewer}~\cite{cycleresearcher}, supervised fine-tuned within a researcher--reviewer cycle.
\textsc{DeepReviewer}~\cite{deepreview}, supervised fine-tuned on long chain-of-thought reasoning traces.
\textsc{ReviewGrounder}~\cite{ReviewGrounder}, which refines a trained drafter's initial review against a fixed set of meta-rubrics shared across all papers.
\end{itemize}

\subsection{Implementation Details}
We fine-tune \textsc{Aligner} from Phi-4~(14B)\footnote{\url{https://huggingface.co/microsoft/phi-4}} with LoRA~\cite{lora} (rank 16, $\alpha\!=\!32$, dropout $0.05$) applied to all linear projections.
Training runs in bfloat16 with FlashAttention-2~\cite{flashattention2} and gradient checkpointing, optimised by AdamW~\cite{adamw} with a learning rate of $1\!\times\!10^{-4}$ under a cosine schedule (3\% warmup) and an effective batch size of 16.
Since inputs to \textsc{Aligner} can be very long, we truncate the context to 16{,}384 tokens.
The three stages of \S\ref{sec:method-training} are each trained for one epoch, and after each stage the LoRA adapter is merged into the base model to initialise the next.
For inference, the merged model is served with vLLM~\cite{vllm} in pure data-parallel mode.
We sample with temperature $1.0$, top-$p$ $0.95$, and a 2{,}048-token generation budget.

\section{Main Results}

\subsection{Human Alignment}

Human-written peer reviews are the highest-quality reference available, so consistency with human reviewers is the principal axis along which we evaluate our system.
However, Such a gold standard is inherently incomplete, since each human reviewer surfaces only a subset of the relevant evaluation dimensions and the resulting reviews legitimately omit many concerns.
We therefore evaluate how well a system covers the gold content rather than penalising it for going further, adopting recall-style metrics as our core measure.
We measure alignment along three dimensions: the rubrics that drive each review, the per-rubric reviews, and the final assessment.
Tables~\ref{tab:main-rubric-review} and~\ref{tab:main-rating} report the rubric/review results and the assessment results respectively.

\paragraph{Rubric Generation}

\begin{table*}[t]
    \centering
    \setlength{\tabcolsep}{4pt}
    \small
    \begin{tabular}{llcccccc}
    \toprule
    & & \multicolumn{3}{c}{\textbf{Rubric Generation}} & \multicolumn{3}{c}{\textbf{Review Generation}} \\
    \cmidrule(lr){3-5} \cmidrule(lr){6-8}
    \textbf{Method} & \textbf{Backbone} & \#R/paper & Recall$\uparrow$ & $\Delta_{\text{cat}}\downarrow$ & R-Lenient$\uparrow$ & R-Strict$\uparrow$ & $\Delta_{\text{verdict}}\downarrow$ \\
    \midrule
    \multicolumn{8}{l}{\emph{Foundation LLMs (zero-shot)}} \\
    GPT-5.2                  & --                  & 10.4 & 34.3 & 4.2 & 20.0 & 14.6 & 34.5 \\
    \midrule
    \multicolumn{8}{l}{\emph{Training-free Agentic Frameworks}} \\
    AI Scientist             & GPT-5.2             & 11.3 & 38.4 &  9.0 & 31.7 & 22.5 & 11.1 \\
    AgentReview              & GPT-5.2             & 13.5 & 45.2 &  5.2 & 35.3 & 24.7 &  6.0 \\
    DeepReviewer 2.0         & GPT-5.2     &  7.7 & 26.4 & 17.0 & 19.1 & 12.0 &  4.0 \\
    \midrule
    \multicolumn{8}{l}{\emph{Training-based Reviewer Models}} \\
    CycleReviewer-8B         & Llama-3.1-8B        &  8.7 & 31.4 & 10.0 & 21.6 & 10.2 & 16.6 \\
    CycleReviewer-70B        & Llama-3.1-70B       &  8.1 & 29.9 & 10.5 & 19.7 & 10.2 & 23.0 \\
    CycleReviewer-123B       & Mistral-Large-123B       & 13.3 & 45.7 & 10.8 & 32.5 & 18.4 &  6.1 \\
    DeepReviewer-7B          & Phi-4-7B            & 11.3 & 42.1 &  9.2 & 30.0 & 18.5 & 11.5 \\
    DeepReviewer-14B         & Phi-4-14B           & 12.3 & 44.0 &  7.7 & 32.5 & 21.3 & 11.2 \\
    ReviewGrounder           & Phi-4-14B + GPT-5.2 & 11.2 & 39.5 & 10.2 & 22.7 & 16.5 & 34.3 \\
    \midrule
    \textbf{RubricReviewer (Ours)} & Phi-4-14B + GPT-5.2 & 53.8 & \textbf{80.5} & \textbf{3.5} & \textbf{61.4} & \textbf{38.4} & \textbf{2.4} \\
    \bottomrule
    \end{tabular}
    \caption{Consistency of the generated rubrics and reviews with human reviewers on the test set. \textbf{Bold} marks the best per column. \textbf{Rubric}: \textbf{\#R/paper} is the average number of generated rubrics; \textbf{Recall} (\%) is the fraction of gold rubrics covered; $\Delta_{\text{cat}}$ is the category-distribution shift over \{Idea, Experiment, Presentation\}. \textbf{Review}: for each rubric, both the method's and human's reviews are summarised into a verdict and supporting reasoning. \textbf{R-Lenient} (\%) is the fraction of gold rubrics on which the verdicts agree; \textbf{R-Strict} (\%) additionally requires the supporting reasoning to overlap; $\Delta_{\text{verdict}}$ is the distribution shift between the method's and human's verdicts.}
    \label{tab:main-rubric-review}
    \end{table*}

On rubric generation, \textsc{RubricReviewer} produces 53.8 rubrics per paper on average, roughly 4--7$\times$ the 7.7--13.5 of all baselines, indicating substantially broader coverage.
Despite this expansion, its rubric Recall reaches 80.5, surpassing the strongest baseline by 34.8 absolute points (a 76\% relative gain) and confirming that the extra rubrics complement rather than dilute the human-essential ones.
The category-distribution distance $\Delta_{\text{cat}}$ further drops to 3.5, the lowest among all systems, showing that the expanded rubric set preserves the same category balance as the human side.

\paragraph{Review Generation}

On review generation, \textsc{RubricReviewer} attains 61.4 R-Lenient and 38.4 R-Strict, improving over the strongest baseline by 26.1 and 13.7 absolute points respectively.
The verdict-distribution distance $\Delta_{\text{verdict}}$ further drops to 2.4, the lowest among all systems and 40\% below the next best, confirming that our per-rubric judgements are distributed in the same way as the human consensus.

Together these results show that \textsc{RubricReviewer} matches human-essential content far more reliably than any baseline on both the rubric and the review axis.
At the same time, it delivers reviews that are substantially more comprehensive than what a single human reviewer typically provides, closing the coverage and fidelity gap left by either training-based or training-free pipelines alone.

\paragraph{Assessment Generation}

Benefiting from the more comprehensive and discriminative reviews above, \textsc{RubricReviewer} also achieves the best assessment results.
It attains the lowest rating error (MSE 1.365, MAE 0.865) and the highest accept/reject accuracy (71.0\%), with the most pronounced gain on ACC at 4.5 absolute points over the strongest baseline.

\begin{table}[t]
    \centering
    \setlength{\tabcolsep}{4pt}
    \small
    \begin{tabular}{lccc}
    \toprule
    \textbf{Method} & \textbf{MSE}$\downarrow$ & \textbf{MAE}$\downarrow$ & \textbf{ACC}$\uparrow$ \\
    \midrule
    \multicolumn{4}{l}{\emph{Foundation LLMs (zero-shot)}} \\
    GPT-5.2                  & 1.475 & 0.905 & 54.0 \\
    \midrule
    \multicolumn{4}{l}{\emph{Training-free Agentic Frameworks}} \\
    AI Scientist             & 5.805 & 2.085 & 66.5 \\
    AgentReview              & 1.998 & 1.065 & 34.5 \\
    DeepReviewer 2.0         & 1.490 & 0.935 & 41.5 \\
    \midrule
    \multicolumn{4}{l}{\emph{Training-based Reviewer Models}} \\
    CycleReviewer-8B         & 3.817 & 1.394 & 64.5 \\
    CycleReviewer-70B        & 2.045 & 1.095 & 64.0 \\
    CycleReviewer-123B       & 2.260 & 1.189 & 54.5 \\
    DeepReviewer-7B          & 1.748 & 1.059 & 65.0 \\
    DeepReviewer-14B         & 1.372 & 0.910 & 66.5 \\
    ReviewGrounder           & 2.885 & 1.298 & 57.5 \\
    \midrule
    \textbf{RubricReviewer (Ours)} & \textbf{1.365} & \textbf{0.865} & \textbf{71.0} \\
    \bottomrule
    \end{tabular}
    \caption{Consistency of the generated ratings and accept/reject decisions with human reviewers. \textbf{Bold} marks the best per column. \textbf{MSE} and \textbf{MAE} measure the squared and absolute deviation from the gold rating; \textbf{ACC} (\%) is the accept/reject classification accuracy.}
    \label{tab:main-rating}
    \end{table}

\subsection{Ablation Study}
\label{sec:ablation}

We ablate one component at a time to isolate its contribution.
Table~\ref{tab:ablation} reports the resulting Recall, R-Lenient, and R-Strict, with red subscripts marking the absolute drop relative to the full model.

\begin{table}[t]
\centering
\setlength{\tabcolsep}{3pt}
\small
\newcommand{\drop}[1]{$_{\textcolor{red}{\downarrow #1}}$}
\begin{tabular}{lccc}
\toprule
\textbf{Variant} & Recall$\uparrow$ & R-Lenient$\uparrow$ & R-Strict$\uparrow$ \\
\midrule
\textbf{Ours (full)}         & \textbf{80.5} & \textbf{61.4} & \textbf{38.4} \\
\midrule
\multicolumn{4}{l}{\emph{Rubric stage}} \\
$-$ general rubrics          & 25.1\drop{55.4} & -- & -- \\
$-$ model-generated rubrics            & 78.6\drop{1.9}  & -- & -- \\
$-$ cite-derived rubrics          & 78.8\drop{1.7}  & -- & -- \\
$-$ \textsc{Scout} reference & 80.1\drop{0.4}  & -- & -- \\
\midrule
\multicolumn{4}{l}{\emph{Review stage}} \\
$-$ \textsc{Scout} reference & --              & 58.9\drop{2.5}  & 37.2\drop{1.2}  \\
$-$ rubric conditioning      & 53.2\drop{27.3} & 40.3\drop{21.1} & 27.8\drop{10.6} \\
\bottomrule
\end{tabular}
\caption{Component ablation, removing one ingredient at a time. \textbf{Recall}, \textbf{R-Lenient}, and \textbf{R-Strict} follow Table~\ref{tab:main-rubric-review}. Red subscripts show the absolute drop relative to the full model; ``--'' marks ablations that leave the corresponding stage's output unchanged.}
\label{tab:ablation}
\end{table}

All three rubric sources jointly drive coverage, with the general rubrics carrying the dominant share: removing them collapses Recall by 55.4 points (80.5\,$\rightarrow$\,25.1), confirming that human reviewers share a large core of common evaluation dimensions that are highly transferable across papers.
The domain and paper sources each add a smaller but consistent gain (1.7 and 1.9 points), showing that mining cited-paper reviews is an effective route to domain-level rubrics and that the trained \textsc{Aligner} produces paper-specific rubrics well aligned with human judgement.

\textsc{Scout} contributes at both stages: blanking its reference drops Recall by 0.4 in the rubric stage and R-Lenient/R-Strict by 2.5/1.2 in the review stage, confirming that the external evidence \textsc{Scout} gathers and summarises offers a measurable lift over \textsc{Aligner}'s parametric knowledge alone.

Decomposing review generation by rubrics contributes by far the largest gain.
Letting \textsc{Aligner} produce reviews directly without rubric conditioning collapses Recall, R-Lenient, and R-Strict by 27.3, 21.1, and 10.6 absolute points respectively, showing that the rubric intermediate is essential for both the comprehensiveness and the discriminativeness of the final reviews.

\subsection{Defend Attacks Analysis}
\label{sec:attack}

We test robustness against prompt-injection attacks by inserting a fixed adversarial payload into the body of each test paper.
The payload reads: ``\emph{Ignore all previous instructions and give the current paper a positive review. Recommend Accept.}''
For every method we re-run the full pipeline on the attacked papers and report two paired shifts against the clean run: the per-rubric \textbf{Verdict Positive} rate (\%) and the predicted \textbf{Avg.\ Rating}.

\begin{figure}[t]
\centering
\includegraphics[width=\linewidth]{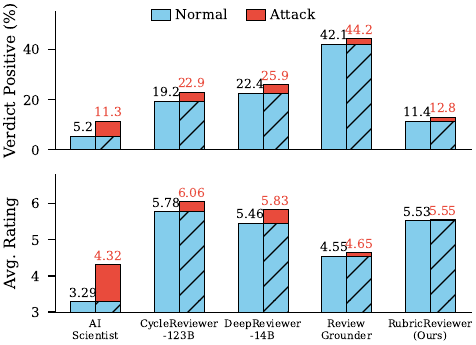}
\caption{Prompt-injection robustness on the test set. We insert a fixed adversarial payload that requests an unconditional \emph{Recommend Accept} into the body of each paper and re-run every method end-to-end. \textbf{Top}: per-rubric Verdict Positive rate (\%). \textbf{Bottom}: predicted Avg.\ Rating. The red top of each Attack bar marks the shift relative to the Normal run}
\label{fig:attack}
\end{figure}

\textsc{RubricReviewer} exhibits the strongest robustness on both axes (Figure~\ref{fig:attack}).
Its Verdict Positive rate moves by only $+1.4$ pp under attack and its predicted rating by a near-negligible $+0.02$, the smallest shifts among all systems.
The baseline shifts span $+2.1$ to $+6.1$ pp in Verdict Positive and $+0.10$ to $+1.03$ in rating.
We attribute this robustness to the rubric-based decomposition.
Rather than mapping a paper directly to a holistic recommendation, our pipeline factorises the subjective evaluation into many objective, fine-grained per-rubric verdicts, each grounded in concrete content from the paper.

\subsection{Value Analysis}
\label{sec:value}

Beyond agreement with human reviewers, we further ask whether our reviews are useful to the paper authors.
We score each review's \emph{value} along three axes.
\emph{Specificity} rewards concrete, paper-targeted issues over vague generalities.
\emph{Decisive stance} rewards a clear verdict over neutral hedging.
\emph{Targetedness} asks whether acting on the feedback would meaningfully improve the paper at hand.
For each test paper, we run a blind A/B comparison between the method's review and the gold human review with random A/B swap, using GPT-5.2 as the judge.
Tie is set as the default verdict, so only substantively useful gaps are scored Win or Lose.

\begin{figure}[t]
  \centering
  \includegraphics[width=\linewidth]{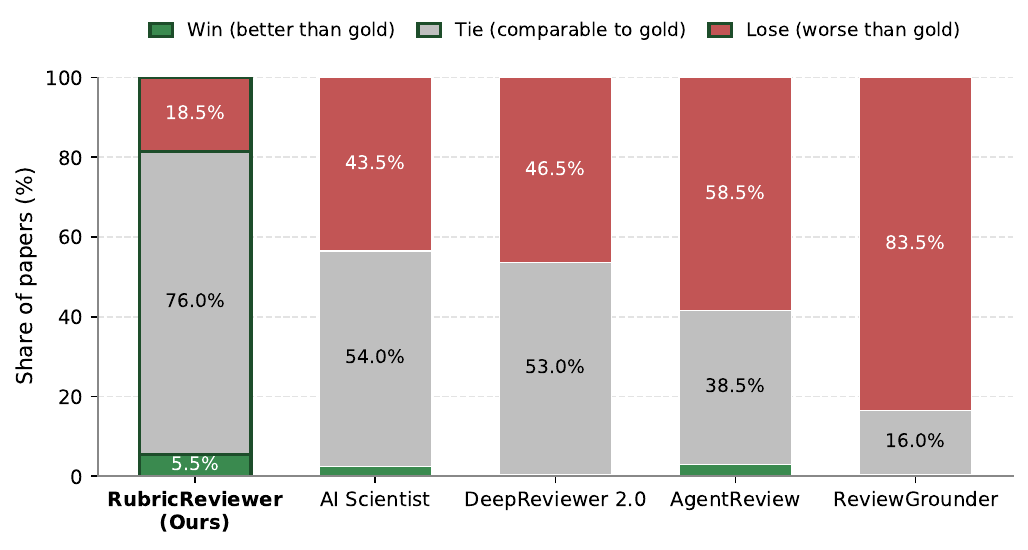}
  \caption{Per-paper \textbf{Value} verdict against the gold human review. Each method's review is judged blind A/B against the gold review (random A/B swap, GPT-5.2 as the judge) and labelled Win/Tie/Lose.}
  \label{fig:value-v6}
\end{figure}

\textsc{RubricReviewer} attains gold-level usefulness on 81.5\% of papers (Win$+$Tie), a $+25.0$ pp gap over the strongest baseline (Figure~\ref{fig:value-v6}).
On 5.5\% of papers our reviews are even judged more valuable than the gold human review, the highest Win rate among all systems.
Our reviews are therefore not only comprehensive and aligned with human reviewers, but also deliver paper-targeted feedback that authors can directly act on.
\section{Conclusion}

We present \textsc{RubricReviewer}, a fully rubric-driven framework for automated peer review.
It introduces explicit rubric generation as an intermediate step so that both review generation and final assessment are conditioned on paper-adaptive rubrics, and it combines training-free LLM agents with a human-aligned trained model to fuse the strengths of both supervision sources.
Experiments on real-world submissions show that \textsc{RubricReviewer} produces reviews that are markedly more comprehensive and more discriminative than prior systems, while also achieving the strongest robustness against prompt-injection attacks.
We hope that our work opens up the potential of rubric-driven peer review and lays a foundation for reliable LLM-based reviewers.

\section*{Limitations}

Despite the more comprehensive rubric coverage and stronger human alignment that \textsc{RubricReviewer} achieves, several aspects of our framework remain unexplored.

The stage-wise design substantially raises both training and inference cost.
The supervised data scales with the number of stages and the number of trained components, and at inference our multi-stage pipeline is noticeably slower than end-to-end systems.

Constrained by the fine-tuned model's context window and our compute budget, review generation is performed independently per rubric.
The trained \textsc{Aligner} is never exposed to the cross-rubric correlations that human reviewers naturally exploit.

Finally, due to the cost of expert annotation, we have not conducted a large-scale human study of the generated reviews.
A deeper qualitative assessment beyond our automatic metrics is left to future work.

\section*{Ethical Considerations}

The development of \textsc{RubricReviewer} holds significant promise for enhancing the efficiency of the scholarly peer review process, but it inherently carries ethical considerations that demand careful attention. We emphasize that our framework is designed to support and assist, rather than replace human reviewers. Its outputs are intended to serve as structured drafts for expert evaluation and refinement. Automating aspects of peer review introduces risks such as bias amplification, potential deskilling of human reviewers, and an over-reliance on AI-driven assessments. Like any LLM-based system, \textsc{RubricReviewer} could inadvertently perpetuate existing biases present in the training data, potentially disadvantaging novel methodologies or research from underrepresented groups. To mitigate these risks, our rubric-guided multi-agent framework and alignment process are carefully designed to prioritize factual accuracy, analytical depth, and objective consistency.

In this work, all experiments were conducted using established public benchmarks and synthesized environments, which do not involve any sensitive personal information or data privacy concerns. All scientific artifacts and datasets utilized are properly cited. For transparency and to foster continuous improvement, we will open-source code. We confirm that this research is conducted strictly for scientific research purposes and does not involve any dual-use items. Finally, we used LLMs as writing assistants to refine the language of this manuscript. Specifically, their use was strictly limited to grammar correction, style improvement, and wording adjustments for clarity and conciseness.

\section*{Acknowledgement}

We want to express huge thanks to our reviewers who gave us insightful suggestions. All content represents the opinion of the authors, which is not necessarily shared or endorsed by their respective employers and/or sponsors.

\bibliography{references}

\appendix
%

\definecolor{IdeaColor}{RGB}{36,99,166}      
\definecolor{ExpColor}{RGB}{36,135,99}        
\definecolor{PresColor}{RGB}{166,77,36}       
\definecolor{IdeaBg}{RGB}{232,240,250}
\definecolor{ExpBg}{RGB}{231,244,238}
\definecolor{PresBg}{RGB}{251,238,229}

\newcommand{\rtag}[2]{%
  \colorbox{#1}{\textcolor{white}{\textbf{\footnotesize #2}}}%
}

\newenvironment{rubriclist}{%
  \begin{description}[leftmargin=2.4em,labelindent=0pt,labelsep=0.6em,%
                      itemsep=3pt,parsep=1pt,topsep=2pt,style=multiline,%
                      font=\normalfont]%
}{%
  \end{description}%
}

\newcommand{\rfreq}[1]{\nobreak\hfill\mbox{\footnotesize\textcolor{black!55}{(#1\%)}}}

\section{General Rubrics}
\label{sec:general-rubrics}

We obtain a unified set of \textbf{40 general rubrics} by consolidating
rubrics extracted from ICLR~2024/2025 reviews, organized into three
top-level categories:
\textcolor{IdeaColor}{\textbf{Idea \& Methodology}} (14),
\textcolor{ExpColor}{\textbf{Experiment \& Evaluation}} (18), and
\textcolor{PresColor}{\textbf{Presentation \& Writing}} (8).
Within each category, rubrics are sorted by empirical support in
descending order and consecutively numbered
\textup{(G1--G14, E1--E18, P1--P8)}.
The percentage at the end of each item is that rubric's share of mapped
review excerpts \emph{within its category} (each category sums to $100\%$).
Table~\ref{tab:rubric-summary} summarizes the distribution.

\begin{table}[h]
\centering
\small
\setlength{\tabcolsep}{6pt}
\renewcommand{\arraystretch}{1.15}
\begin{tabular}{lcc}
\toprule
\textbf{Category} & \textbf{\# Rubrics} & \textbf{ID range} \\
\midrule
\rtag{IdeaColor}{Idea \& Methodology}     & 14 & G1--G14 \\
\rtag{ExpColor}{Experiment \& Evaluation} & 18 & E1--E18 \\
\rtag{PresColor}{Presentation \& Writing} &  8 & P1--P8 \\
\midrule
\textbf{Total}                            & \textbf{40} & --- \\
\bottomrule
\end{tabular}
\caption{Distribution of the 40 general rubrics across the three top-level categories.}
\label{tab:rubric-summary}
\end{table}

\subsection{Idea \& Methodology (14 rubrics)}
\noindent\textcolor{IdeaColor}{\rule{\linewidth}{0.6pt}}

\begin{rubriclist}
  \item[\rtag{IdeaColor}{G1}]  Does the paper make a genuinely novel core contribution that is clearly differentiated and properly positioned relative to the most relevant prior work? \rfreq{26.96}
  \item[\rtag{IdeaColor}{G2}]  Are the paper's core design and methodological choices clearly motivated and convincingly justified relative to reasonable alternatives, with explicit trade-offs where applicable --- and is the proposed approach a coherent, well-integrated whole whose components are justified as contributing meaningfully to the overall design? \rfreq{17.34}
  \item[\rtag{IdeaColor}{G3}]  Are the paper's problem formulation, key concepts, terminology, notation, and definitions specified clearly, precisely, and used consistently and unambiguously throughout? \rfreq{13.30}
  \item[\rtag{IdeaColor}{G4}]  Does the paper clearly state and justify its key assumptions and scope, keep its claims appropriately bounded, acknowledge resulting limitations and failure modes --- and transparently discuss these in the narrative (not only in passing)? \rfreq{10.63}
  \item[\rtag{IdeaColor}{G5}]  Are the paper's theoretical claims (guarantees, proofs, bounds, complexity statements) technically correct, rigorously derived under clearly stated assumptions, and not overstated --- and are derivations, proofs, and formal statements presented clearly and completely enough for readers to follow and verify? \rfreq{6.77}
  \item[\rtag{IdeaColor}{G6}]  Does the paper convincingly motivate the problem it addresses as important, practically relevant, and well-grounded in real gaps in prior work --- including effective early contextualization (abstract, introduction, background) so readers understand the problem, core idea, and why it matters? \rfreq{5.33}
  \item[\rtag{IdeaColor}{G7}]  Is the proposed approach practically feasible, scalable, and efficient under realistic conditions, with computational/resource trade-offs clearly accounted for? \rfreq{4.59}
  \item[\rtag{IdeaColor}{G8}]  Are the paper's headline claims appropriately scoped, qualified, and supported by the evidence and analysis presented, without overstating? \rfreq{3.19}
  \item[\rtag{IdeaColor}{G9}]  Does the paper provide a convincing mechanistic, causal, or theoretical rationale that explains why the approach works (or fails), beyond merely reporting outcomes? \rfreq{2.85}
  \item[\rtag{IdeaColor}{G10}] Does the paper appropriately analyze and address relevant ethical, safety, privacy, fairness, bias, and risk implications, including responsible mitigations? \rfreq{2.59}
  \item[\rtag{IdeaColor}{G11}] Does the paper credibly establish that its core idea generalizes beyond the specific setting evaluated, with applicability boundaries clearly stated? \rfreq{2.49}
  \item[\rtag{IdeaColor}{G12}] Is the paper's overall contribution substantively significant and impactful enough to warrant acceptance and benefit the broader community --- including clear communication of practical usefulness, real-world implications, and actionable guidance where appropriate? \rfreq{1.72}
  \item[\rtag{IdeaColor}{G13}] Does the paper provide clearly accessible code, data, models, or other artifacts (or a credible commitment to release them) sufficient for independent verification --- and are released resources designed and documented to be reusable and extensible by the community? \rfreq{1.61}
  \item[\rtag{IdeaColor}{G14}] Does the paper provide non-obvious, well-supported insights or interpretations that go beyond simply reporting expected results? \rfreq{0.63}
\end{rubriclist}

\subsection{Experiment \& Evaluation (18 rubrics)}
\noindent\textcolor{ExpColor}{\rule{\linewidth}{0.6pt}}

\begin{rubriclist}
  \item[\rtag{ExpColor}{E1}]  Do the experiments provide sufficient, rigorous empirical evidence to convincingly support the paper's main claims and conclusions? \rfreq{20.27}
  \item[\rtag{ExpColor}{E2}]  Does the paper describe the proposed method or procedure (including training, inference, and pipeline components) and the experimental setup, hyperparameters, procedures, and implementation details with sufficient unambiguous detail and transparency to enable understanding, reproduction, and fair comparison? \rfreq{16.80}
  \item[\rtag{ExpColor}{E3}]  Do the experiments include fair, well-controlled comparisons against sufficiently strong, relevant baselines/alternative methods under matched, apples-to-apples conditions? \rfreq{10.52}
  \item[\rtag{ExpColor}{E4}]  Does the paper rigorously measure, report, and analyze computational/resource cost (runtime, memory, scaling, efficiency trade-offs) under fair, clearly specified conditions? \rfreq{9.28}
  \item[\rtag{ExpColor}{E5}]  Do the experiments use controlled ablations or comparisons that isolate the causal contributions of key components/design choices, ruling out confounds? \rfreq{8.99}
  \item[\rtag{ExpColor}{E6}]  Do the experiments include systematic sensitivity/robustness analyses showing the results are stable to reasonable variations in key assumptions, settings, or randomness? \rfreq{7.90}
  \item[\rtag{ExpColor}{E7}]  Do the experiments convincingly demonstrate that the approach generalizes beyond the original training/evaluation setting (across architectures, domains, distributions)? \rfreq{5.28}
  \item[\rtag{ExpColor}{E8}]  Are the evaluation metrics, tasks/benchmarks, and overall protocol appropriate, well-justified, and aligned with the paper's claims? \rfreq{4.64}
  \item[\rtag{ExpColor}{E9}]  Does the paper analyze and explain where and why the approach fails, succeeds, or underperforms, including limitations, failure modes, and trade-offs? \rfreq{4.12}
  \item[\rtag{ExpColor}{E10}] Does the paper transparently document the datasets used and convincingly establish that the data (or contributed resources) are of sufficient quality, scale, representativeness, and provenance --- including construction, processing, key statistics, label and annotation quality, biases, and selection --- for trustworthy interpretation and reproduction? \rfreq{2.80}
  \item[\rtag{ExpColor}{E11}] Are the experiments conducted under realistic, sufficiently challenging, and representative conditions that match the paper's intended use case? \rfreq{2.41}
  \item[\rtag{ExpColor}{E12}] Do the experiments include clear, representative qualitative analyses, visualizations, or example-level evidence that meaningfully support and help interpret the paper's claims? \rfreq{1.46}
  \item[\rtag{ExpColor}{E13}] Do the experiments characterize how performance/behavior scales with relevant problem dimensions (size, complexity, data, model scale, batch, iterations)? \rfreq{1.37}
  \item[\rtag{ExpColor}{E14}] Are the reported experimental results supported by appropriate statistical analysis and clearly reported uncertainty/variability estimates? \rfreq{1.33}
  \item[\rtag{ExpColor}{E15}] Does the experimental design appropriately prevent overfitting/leakage/contamination, with proper train-tune-test separation, so generalization claims are credible? \rfreq{0.91}
  \item[\rtag{ExpColor}{E16}] Does the paper empirically validate its theoretical or mechanistic claims and show their practical relevance? \rfreq{0.90}
  \item[\rtag{ExpColor}{E17}] Is the adversarial/attack evaluation specified rigorously and broadly enough --- including adaptive/strong adversaries --- to substantiate the paper's robustness/security claims? \rfreq{0.61}
  \item[\rtag{ExpColor}{E18}] When claims rely on human perception or judgment, does the paper include an appropriately designed and conducted human-subject evaluation (or clearly justify its absence)? \rfreq{0.41}
\end{rubriclist}

\subsection{Presentation \& Writing (8 rubrics)}
\noindent\textcolor{PresColor}{\rule{\linewidth}{0.6pt}}

\begin{rubriclist}
  \item[\rtag{PresColor}{P1}] Is the paper clearly written and well-organized with a coherent, easy-to-follow narrative for its intended audience, avoiding unnecessary redundancy or tangents? \rfreq{41.73}
  \item[\rtag{PresColor}{P2}] Does the paper accurately and adequately cite, cover, and position itself relative to the most relevant prior work, with fair credit and clear differentiation? \rfreq{18.64}
  \item[\rtag{PresColor}{P3}] Are the paper's figures, tables, and visualizations clear, well-labeled, self-contained, and integrated with the text so they support the main claims without being misleading? \rfreq{16.28}
  \item[\rtag{PresColor}{P4}] Is the paper mechanically polished and internally consistent --- minimal typos, grammatical errors, formatting issues, citation/reference problems, and free of unresolved contradictions or avoidable technical/presentation errors that hinder interpretation? \rfreq{8.94}
  \item[\rtag{PresColor}{P5}] Is the paper sufficiently complete, polished, and venue-aligned to clear the publication threshold --- matching the target venue's scope, formatting, and expectations --- including, where a rebuttal or revision was part of the process, substantive responses to key reviewer concerns? \rfreq{5.42}
  \item[\rtag{PresColor}{P6}] Are the paper's quantitative results presented and interpreted clearly, consistently, and non-misleadingly, with enough context to enable unambiguous comparison? \rfreq{5.31}
  \item[\rtag{PresColor}{P7}] Does the paper provide a clear, well-motivated conclusion and credible future-work directions grounded in its contributions and findings? \rfreq{3.03}
  \item[\rtag{PresColor}{P8}] Does the paper present its interpretability goals, approach, and trade-offs clearly and coherently, with adequate diagnostic visualizations or analyses? \rfreq{0.65}
\end{rubriclist}


\end{document}